\documentclass{article}

     \usepackage[preprint]{neurips_2020}

\usepackage[utf8]{inputenc} 
\usepackage[T1]{fontenc}    
\usepackage{hyperref}       
\usepackage{url}            
\usepackage{booktabs}       
\usepackage{amsfonts}       
\usepackage{nicefrac}       
\usepackage{microtype}      
\usepackage{graphicx} 
\usepackage[labelfont=bf]{caption}
\title{Large Language Models for Granularized Barrett’s Esophagus Diagnosis Classification}

\author{%
  Jenna~Kefeli\thanks{Co-first authors} \\
  Department of Systems Biology\\
  Columbia University\\
   \And
     Ali~Soroush\footnotemark[1]\\
   Division of Digestive and Liver Diseases\\ Department of Medicine \\
  Columbia University Irving Medical Center\\
   \And
     Courtney J.~Diamond\\
   Department of Biomedical Informatics\\
  Columbia University \\
   \And
     Haley M.~Zylberberg\\
   Department of Biomedical Informatics\\
  Columbia University \\
   \And
     Benjamin~May\\
   Herbert Irving Comprehensive Cancer Center\\
    Columbia University Irving Medical Center\\
   \And
     Julian A.~Abrams\\
   Division of Digestive and Liver Diseases \\Department of Medicine \\
    Columbia University Irving Medical Center\\
   \And
     Chunhua~Weng\\
   Department of Biomedical Informatics\\
  Columbia University \\
   \And
     Nicholas~Tatonetti\thanks{Corresponding author: nicholas.tatonetti@cshs.org} \\
  Department of Computational Biomedicine\\
  Cedars-Sinai \\
}

\begin{document}

\maketitle

\begin{abstract}
  Diagnostic codes for Barrett’s esophagus (BE), a precursor to esophageal cancer, lack granularity and precision for many research or clinical use cases. Laborious manual chart review is required to extract key diagnostic phenotypes from BE pathology reports. We developed a generalizable transformer-based method to automate data extraction. Using pathology reports from Columbia University Irving Medical Center with gastroenterologist-annotated targets, we performed binary dysplasia classification as well as granularized multi-class BE-related diagnosis classification. We utilized two clinically pre-trained large language models, with best model performance comparable to a highly tailored rule-based system developed using the same data. Binary dysplasia extraction achieves 0.964 F1-score, while the multi-class model achieves 0.911 F1-score. Our method is generalizable and faster to implement as compared to a tailored rule-based approach. 
\end{abstract}

\section{Introduction}

Barrett’s esophagus (BE) is a premalignant transformation of the esophageal lining and precursor to esophageal cancer, estimated to impact 1.5\% of the overall population [1]. Patients diagnosed with BE are routinely monitored for any progression in severity, as measured by degree of tissue abnormality (i.e., dysplasia). The full spectrum of BE dysplasia categories is not fully captured in the EHR by ICD codes. Currently, laborious manual chart review is required to classify upper endoscopy pathology reports into appropriate BE diagnoses with sufficient accuracy [2]. Automating this process can increase the efficiency of extracting BE outcome measures for research and clinical use [3]. 

The advent of large language models (LLMs) has led to significant advances in natural language processing (NLP), enabling automated methods with reliable performance and broad extensibility [4-6]. However, no studies have demonstrated the use of LLMs for BE-related NLP use cases. One previous study used a hybrid machine learning-rule-based NLP approach to classify reports with binary dysplasia prediction, but did not extend prediction to multi-class BE diagnosis [7].

In this study, we apply two BERT-based large language models to perform automated BE diagnosis classification, using patient pathology reports from Columbia University Irving Medical Center (CUIMC). We compare our method to a rule-based method developed previously at CUIMC, which used regular expressions and Metamap for BE diagnosis classification [8]. As the rule-based method was developed based on the same data used in this study, it serves as a direct benchmark.

\section{Results}

\subsection{Characterization of Barrett’s esophagus (BE) Pathology Report Dataset }

All endoscopy pathology reports from patients who underwent surveillance for BE between January 1, 2016 and December 31, 2020 were obtained from the CUIMC data warehouse. Patients were randomly assigned to either the development or validation set, with 150 patients in each. Patients were excluded if they did not meet the age criterion or did not meet diagnostic criteria for BE (Figure 1, see Methods). Demographic characteristics of the development and validation sets are presented in Table 1. 

Gastroenterologists manually annotated report-level BE-related diagnoses to generate a gold standard. One gastroenterologist manually annotated the development set, and two gastroenterologists annotated the validation set (with a third gastroenterologist serving as tie-breaker). The final dataset contained 619 pathology reports corresponding to 214 patients. BE diagnosis class distributions for both the development and validation sets are shown in Table 2. We observe that BE diagnosis classes are mostly similarly distributed across the two sets, except for a slight imbalance in the proportion of reports annotated as BE with low-grade dysplasia and reports annotated as esophageal adenocarcinoma.

\subsection{Fine-Tuning of Clinically Trained Large Language Models}

For model training, the development set (n=301 reports) was divided into training and evaluation sets, which were used for hyperparameter optimization (see Methods). The training set consisted of 240 reports, and the evaluation set contained 61 reports. We fine-tuned pre-trained BERT-based [9] LLMs on both full and trimmed (diagnosis-related sub-sections) pathology reports. Hyperparameter optimization was performed over model type, maximum input token length, and learning rate (see Methods). 

We included both ClinicalBERT [10] and Clinical-BigBird [11] models in the hyperparameter space. A more recently developed model, Clinical-BigBird has a longer input capacity of 4,096 tokens (as compared to ClinicalBERT’s maximum input of 512 tokens). Clinical-BigBird allowed the input of all text in this dataset for training, especially important as >20\% of full-text reports had >512 tokens per report (Supplementary Table 1). 

After training, the best model was chosen based on the evaluation set, and then applied to the held-out validation set for independent testing. Two separate classification tasks were performed: binary dysplasia diagnosis classification (no dysplasia versus dysplasia or worse) and multi-class BE diagnosis classification. For each task, we trained and tested our models on both the sub-sectioned reports and minimally processed full-report datasets separately. We compared our method to a previously developed rules-based approach, which had been developed and tested on exclusively on sub-sectioned reports [8].

\subsection{Binary Dysplasia Report Classification}

We first fine-tuned models to predict the binary target. In the development set, approximately 20\% of reports contained a diagnosis of dysplasia (or worse), as did 18\% of validation set reports (Table 2). We report best-model performance on held-out validation report text (Table 3, Supplementary Table 2). Overall, development set performance was higher than validation set performance, as expected (Supplementary Table 2). Validation set performance for the sub-section-trained BERT-based model was slightly higher than that of the full-report-trained BERT-based model (Table 3), with sub-section accuracy 97.8\% compared to 96.5\% for full reports. Best-performing BERT-based models performed similarly to the customized, rule-based method8 across almost all metrics (Table 3). 

\subsection{Multi-Class BE Diagnosis Report Classification}

To further refine our prediction target and test a higher-complexity task, we performed multi-class classification across 6 BE diagnosis classes. For this task, the dataset exhibited greater imbalance amongst classes as compared to the binary task, with 4 of 6 diagnosis categories having <10\% representation in the development set (Table 2). We report best-model performance on held-out validation report text (Table 4, Supplementary Table 3). All evaluation metrics were macro-calculated for the multi-class task. Development set performance was higher than validation set performance, as expected (Supplementary Table 3). Validation set performance for the full-report-trained BERT-based model is better than performance for the sub-section-trained BERT-based model across all metrics (Table 4). Best-performing BERT-based models did not perform as well as the customized rule-based method [8], for both sub-section-trained and full-report-trained models. However, the full-report-trained model performance approached that of the rule-based method, achieving an F1-score of 0.911 and accuracy of 92.1\% (compared to F1-score of 0.958 and 97.\% accuracy of for the rule-based method).

\section{Discussion}

In this study, we trained and tested BERT-based large language models using sub-section and full-report text inputs for binary dysplasia and multi-class BE diagnosis classification. We compared our results to those of a rule-based method developed previously specifically for this pathology report dataset. We found that our best-model performance was overall comparable to that of a previously developed rule-based method [8]. The rule-based method exhibited slightly higher performance, but required very time-intensive and expertise-dependent iterative rule development, customized for this specific dataset. Importantly, our model pipeline was substantially faster to develop and implement. Other institutions, particularly ones that are resource-limited, could benefit from implementing a similar NLP development system. Automating BE diagnosis classification could shorten time for chart review, research cohort construction, and generate datasets for downstream multi-modal research studies [3,12].

We found that training with full reports led to models with similar performance compared to using sub-sectioned reports, suggesting that our method does not require report pre-processing during preparation for model input. This is useful for generalizability. Further, we found that in three of four sub-tasks, Clinical-BigBird (with 2,048 maximum input tokens) outperformed ClinicalBERT (see Methods). As mentioned above, ClinicalBERT is currently one of the most frequently used models for clinical NLP research. Based on our results, we would recommend that researchers who are already utilizing ClinicalBERT might also try implementing Clinical-BigBird, as they may find better performance on clinical tasks generally. Both ClinicalBERT and Clinical-BigBird have similar implementation requirements, although Clinical-BigBird consumes more memory at run-time. 

A limitation of our method is that protected health information (PHI) was used for training. To publicly release our trained model for external use, we would be required to implement privacy-focused safeguards [13] to prevent any possible de-identification. It is therefore more efficient for individual institutions to develop internal BERT-based NLP pipelines similar to ours in order to automate BE diagnosis report classification. We have made our code available for this purpose. As seen in our study, implementation of this method should be feasible even if only smaller-sized annotated datasets are available. 

In a similar study, Nguyen Wenker et al. performed binary dysplasia classification, performing slightly better across evaluation metrics as compared with our method [7]. However, their work differed from ours in a number of aspects, requiring almost double the number of annotated training reports and utilizing a user-defined iterative approach (similar to [8]), requiring greater expertise and development time. In comparison, our approach extends to both binary and multi-class dysplasia classification, utilizes a straightforward and directly applicable large language model with fewer training input samples, and requires shorter development time. 

In future work, we may further validate our model by evaluating on CUIMC pathology reports from time periods not covered in this study and assess external validity using pathology reports from outside medical systems. We may also compare our method to few-shot or zero-shot classification approaches, using state-of-the-art LLMs like PaLM 2 [14] or GPT-4 [15], to further reduce the development time of automated clinical NLP applications. 

\section{Figures and Tables}

\begin{figure}[ht]
  \centering
  \includegraphics[width=.7\linewidth]{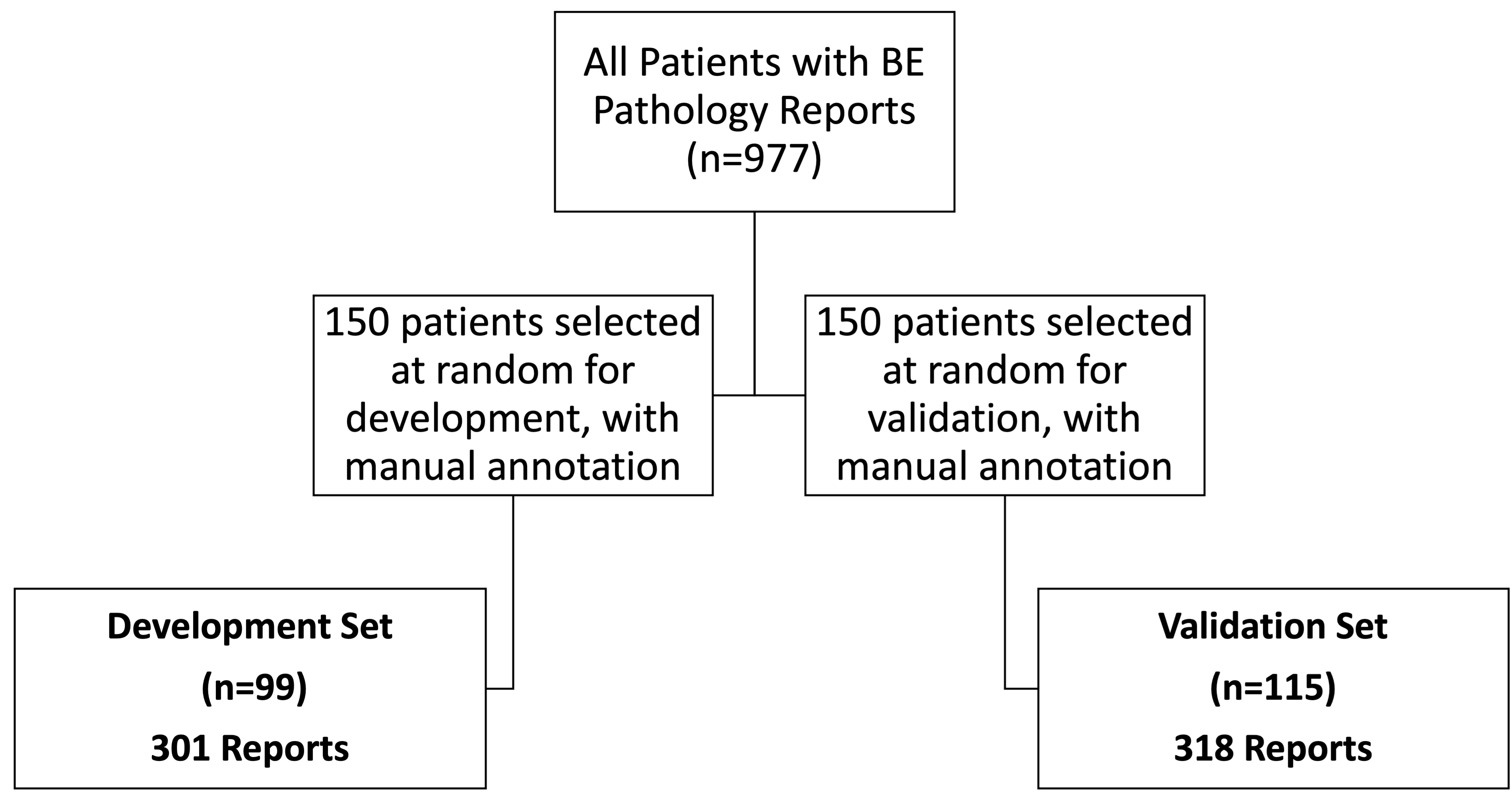} %
  \caption{Data selection for CUIMC cohort. Final selected reports in both the development and validation sets were manually annotated by gastroenterologists.}
\end{figure}

\begin{table}[ht]
  \caption{Demographic Characteristics of CUIMC Cohort. }
  \label{table1}
  \centering
  \begin{tabular}{ccc}
    \toprule
{\bf Patient Characteristics} & {\bf Development} (n=99) & {\bf Validation} (n=115)\\
    \midrule
{\bf Age (years), avg +/- std }  & 68.8 +/- 8.5  & 68.6 +/- 8.3\\
 \midrule
{\bf Sex, n (\%) }& & \\  
Male & 71 (74.0\%) & 87 (77.7\%) \\
Female  & 25 (26.0\%)  & 25 (22.2\%) \\
\midrule
{\bf Race, n (\%) } & &\\ 
White & 88 (91.7\%) & 95 (84.8\%)\\
Non-White & 5 (5.2\%) & 10 (8.9\%) \\
Other/Unknown/Declined  & 3 (3.1\%) &   7 (6.2\%)\\
\midrule
{\bf Ethnicity, n (\%) } & &\\ 
Not Hispanic  & 82 (85.4\%) &   86 (76.8\%) \\
Hispanic & 2 (2.1\%) & 2 (1.8\%) \\
Unknown/Declined & 12 (12.5\%) &   24 (21.4\%)\\
\midrule
{\bf Median pathology reports per patient (IQR) } &  2.0 (1.0-4.0) & 2.0 (1.0-4.0)\\
    \bottomrule
  \end{tabular}
\end{table}

\begin{table}[ht]
   \caption{Report-based distribution of BE diagnosis across development and validation sets. Esophageal adenocarcinoma (EAC), high-grade dysplasia (HGD), low-grade dysplasia (LGD).}
    \label{table2}
  \centering
  \begin{tabular}{lll}
    \toprule
    {\bf BE Diagnosis}    & {\bf Development Set, n (\%)}    & {\bf Validation Set, n (\%)}\\
    \midrule
EAC & 11 (3.7\%) & 22 (6.9\%) \\
BE with HGD & 30 (10.0\%) & 26 (8.2\%) \\
BE with LGD & 18 (6.0\%) & 8 (2.5\%) \\
BE indefinite for dysplasia & 15 (5.0\%) & 22 (6.9\%) \\
BE with no dysplasia & 88  (29.2\%) & 99 (31.1\%) \\
No histological evidence of BE &  139 (46.2\%) & 141 (44.3\%) \\
    \bottomrule
  \end{tabular}
\end{table}

\begin{table}[ht]
  \caption{Binary dysplasia classification, validation set performance.}
  \label{table3}
  \centering
  \begin{tabular}{ccccccc}
    \toprule
    {\bf Report Type}  & {\bf Model}  & {\bf Recall}  &  {\bf Precision}  & {\bf Accuracy}  & {\bf F1-Score}  & {\bf AU-ROC} \\
    \midrule
Sub-sectioned & BERT-Based & 1.000 & 0.889 & 0.978 & 0.964 & 0.997 \\
Sub-sectioned & Rule-Based & 1.000 & 0.966 & 0.990 & 0.982 & - \\
Full &  BERT-Based & 0.911 & 0.895 & 0.965 & 0.941 & 0.985 \\
    \bottomrule
  \end{tabular}
\end{table}

\begin{table}[ht]
  \caption{Multi-class BE diagnosis prediction, validation set performance.}
  \label{table4}
  \centering
  \begin{tabular}{ccccccc}
    \toprule
    {\bf Report Type}  & {\bf Model}  & {\bf Recall}  &  {\bf Precision}  & {\bf Accuracy}  & {\bf F1-Score}  & {\bf AU-ROC} \\
    \midrule
Sub-Sectioned & BERT-Based & 0.878 & 0.893 & 0.915 & 0.876 & 0.989 \\
Sub-Sectioned & Rule-Based& 0.973 & 0.946 & 0.975 & 0.958 & - \\
Full & BERT-Based & 0.939 & 0.898 & 0.921 & 0.911 & 0.990 \\
    \bottomrule
  \end{tabular}
\end{table}

\section{Data Availability Statement.} Pathology reports are not available for public dissemination due to PHI. Code for this study may be found in the following repository: \begin{verbatim} www.github.com/jkefeli/BE_BERT\end{verbatim}

\section*{References}

\medskip

\small

\noindent [1]  Eluri S, Shaheen NJ. Barrett’s esophagus: Diagnosis and management. {\it Gastrointestinal Endoscopy}. 2017;85(5):889-903. 

\noindent [2]  Zozus MN, Pieper C, Johnson CM, et al. Factors Affecting Accuracy of Data Abstracted from Medical Records. {\it PLoS One}. 2015;10(10):e0138649.

\noindent [3]  Juhn Y, Liu H. Artificial intelligence approaches using natural language processing to advance EHR-based clinical research. {\it J Allergy Clin Immunol.} 2020;145(2):463-469.

\noindent [4]  Wu S, Roberts K, Datta S, et al. Deep learning in clinical natural language processing: a methodical review. {\it J Am Med Inform Assoc. }2020;27(3):457-470.

\noindent [5]  Chen A, Yu Z, Yang X, Guo Y, Bian J, Wu Y. Contextualized medication information extraction using Transformer-based deep learning architectures. {\it J Biomed Inform.} 2023;142:104370.

\noindent [6]  Yang X, Chen A, PourNejatian N, et al. A large language model for electronic health records. {\it NPJ Digit Med.} 2022;5(1):194.

\noindent [7]  Nguyen Wenker T, Natarajan Y, Caskey K, et al. Using Natural Language Processing to Automatically Identify Dysplasia in Pathology Reports for Patients With Barrett's Esophagus. {\it Clin Gastroenterol Hepatol.} 2023;21(5):1198-1204. 

\noindent [8]  Soroush A, Diamond CJ, Zylberberg HM, et al. Natural Language Processing Can Automate Extraction of Barrett's Esophagus Endoscopy Quality Metrics. Preprint. {\it medRxiv.} 2023;2023.07.11.23292529. 

\noindent [9]  Vaswani, A, et al. Attention is all you need. {\it Advances in neural information processing systems.} 2017.

\noindent [10] Alsentzer, E, et al. Publicly available clinical BERT embeddings. {\it arXiv.} 2019.

\noindent [11] Li Y, Wehbe RM, Ahmad FS, Wang H, Luo Y. A comparative study of pretrained language models for long clinical text. {\it J Am Med Inform Assoc. }2023;30(2):340-347.

\noindent [12] Acosta JN, Falcone GJ, Rajpurkar P, Topol EJ. Multimodal biomedical AI. {\it Nat Med.} 2022;28(9):1773-1784.

\noindent [13] Ficek J, Wang W, Chen H, Dagne G, Daley E. Differential privacy in health research: A scoping review. {\it J Am Med Inform Assoc.} 2021;28(10):2269-2276.

\noindent [14] Anil, R, et al. PaLM 2 Technical Report. {\it arXiv.} 2023.

\noindent [15] OpenAI. GPT-4 technical report. {\it arXiv.} 2023.

\noindent [16] Wolf, T, et al. Huggingface's transformers: State-of-the-art natural language processing. {\it arXiv.} 2019.

\section{Methods}

\subsection{Report Pre-Processing and Annotation} 

Full reports were extracted from the CUIMC data warehouse and minimally pre-processed to remove extraneous punctuation marks generated during data extraction. Sub-sectioned reports were additionally processed to only include sections of the pathology report with diagnostic information. Sub-sectioned reports did not include information such as patient name, age, relevant clinical history, and pathologist signatures. 

During annotation, patients were labeled as not having BE if they did not have at least one endoscopy with at least 1 cm of salmon-colored mucosa and a finding of intestinal metaplasia present in biopsies of said mucosa.

\subsection{Training and Optimization Specifications } 

The dataset was split into development/validation sets at the patient-level, and the development set was then split into training/evaluation sets at the report-level. The development set was used for hyperparameter optimization to ascertain the best-performing model per task, whereas the validation set was held-out and used for final evaluation only. 

We included both ClinicalBERT and Clinical-BigBird as models, both of which were previously trained on the same set of publicly available clinical notes. These models differ in architecture and maximum input report length, i.e., maximum number of tokens used for model input in a single report. ClinicalBERT is currently widely used in NLP research as a standard pretrained model for fine-tuning clinical tasks, while Clinical-BigBird is a more recently developed model and less commonly adopted. 

For hyperparameter optimization, we performed a grid search across two learning rates, three random seeds, two model types, and various maximum input token size (512 tokens for ClinicalBERT; 512, 1,024, and 2,048 tokens for Clinical-BigBird). We also varied batch size according to maximum input token size, as necessitated by the memory constraints of our hardware system. We used the Hugging Face transformers package [16] for model implementation. Each model instance took approximately 55 minutes to run. Across the hyperparameter space, all models were trained in less than one day per task. 

The following metrics were used for model performance evaluation: Recall, precision, accuracy, F1-score, and AU-ROC. All evaluation metrics are macro-computed for the multi-class task. For the models trained on full reports, we additionally evaluated F2-Beta. For hyperparameter optimization, both AU-ROC and F2-Beta were tested as optimization metrics. 

For comparison, we included the performance of a rule-based method developed based on sub-sectioned CUIMC upper endoscopy reports using the same development/validation report split [8]. The rule-based study did not assess performance using AU-ROC and F2-Beta.

\subsection{Binary Dysplasia Task } 

For the binary dysplasia diagnosis classification task, we defined the prediction target as either 1 (dysplasia or worse – i.e., EAC, BE with HGD, or BE with LGD, as in Table 2), or 0 (no dysplasia – i.e., BE without dysplasia, BE indefinite for dysplasia, or no histological evidence of BE). We ran optimization experiments as outlined above for the hyperparameter grid search, on both the sub-section and the full-report datasets. We considered the “best performing model” to be the model that had the highest optimization evaluation metric on the development set. We found that the best-performing model for sub-sections was Clinical-BigBird (with 2048 input tokens), whereas the best-performing model for full reports was ClinicalBERT (with 512 input tokens). 

\subsection{Multi-Class Diagnosis Task }

For the multi-class BE diagnosis classification task, we included all six diagnosis classes as defined in Table 2. We ran optimization experiments as outlined above for the hyperparameter grid search, on both the sub-sectioned report and the full report datasets. We considered the “best performing model” to be the model that had the highest optimization evaluation metric on the development set. We found that the best-performing model for both the sub-sectioned report and full report datasets was Clinical-BigBird (with 2048 input tokens). 

\newpage
\section{Supplementary Data}
\begin{table}[ht]
  \caption*{{\bf Supplementary Table 1.} Per-Report Tokenization Statistics. Min, max, and percentile (p) number of tokens per report. \%>512 is percent of reports with greater than 512 tokens per report.}
  \label{supplementary_table_1_a}
  \centering
  \begin{tabular}{lllllll}
  \end{tabular}
\end{table}

\begin{table}[ht]
\small
  \caption*{{\bf Supplementary Table 1A.} Development set.}
  \label{supplementary_table_1_a}
  \centering
  \begin{tabular}{lllllll}
    \toprule
    {\bf Tokenizer}  & Min & 25p &50p& 75p& Max & \%  > 512 \\ 
 \midrule
 ClinicalBERT   &  &&&&&\\        
Sub-Section Text & 33 & 110 &189 &289 &1179 & 4.7 \\
Full Report Text & 57 & 239 &356& 532& 1782 & 26.2\\
 \midrule
Clinical-BigBird  &&&&&\\        
Sub-Section Text & 32 & 102 &168 &252& 1116 & 3.7\\
Full Report Text & 60 & 221 &325& 482 &1641 & 20.9\\
    \bottomrule
  \end{tabular}
\end{table}

\begin{table}[ht]
\small
  \caption*{{\bf Supplementary Table 1B.} Validation set.}
  \label{supplementary_table_1_b}
  \centering
  \begin{tabular}{lllllll}
    \toprule
    {\bf Tokenizer}  & Min & 25p &50p& 75p& Max & \%  > 512 \\ 
 \midrule
 ClinicalBERT   &  &&&&&\\        
Sub-Section Text & 44 & 109 &187 &278 &1312 & 7.5\\
Full Report Text & 64 & 238 &343 &569 &1720 & 29.6\\
 \midrule
Clinical-BigBird  &&&&&\\        
Sub-Section Text & 41  &99  &167 &246 &1238 & 4.7\\
Full Report Text & 62  &223 &316 &524 &1520 & 26.1\\
    \bottomrule
  \end{tabular}
\end{table}

\newpage
\begin{table}[ht]
\small
  \caption*{{\bf Supplementary Table 2.} Expanded results for binary dysplasia classification.}
  \label{supplementary_table_2}
  \begin{tabular}{lllllllll}
    \toprule
    {\bf Report Type}  & {\bf Model} & {\bf Dataset} &  {\bf AU-ROC} & {\bf Recall}  &  {\bf Precision}  & {\bf Accuracy}  & {\bf F1-Score}  & {\bf F2-
Beta}\\
    \midrule
Sub-Section & BERT-Based & Development & 1.000 &0.949 &1.000& 0.990 &0.984 &- \\
   & & Validation & 0.997 &1.000 &0.889 &0.978 &0.964& - \\
Sub-Section &Rule-Based & Development& -& 1.000 &0.938 &0.989 &0.968 &- \\
   &&  Validation & - &1.000& 0.966 &0.990& 0.982 &- \\
Full & BERT-Based  &Development& 0.997& 0.983 &0.983 &0.993 &0.990 &0.990 \\
   & & Validation & 0.985 &0.911& 0.895& 0.965 &0.941 &0.943 \\
    \bottomrule
  \end{tabular}
\end{table}

\begin{table}[ht]
\small
  \caption*{{\bf Supplementary Table 3.} Expanded results for multi-class BE diagnosis classification.}
  \label{supplementary_table_3}
  \begin{tabular}{lllllllll}
    \toprule
 {\bf Report Type}  & {\bf Model} & {\bf Dataset} &  {\bf AU-ROC} & {\bf Recall}  &  {\bf Precision}  & {\bf Accuracy}  & {\bf F1-Score}  & {\bf F2-
Beta}\\
    \midrule
Sub-Section &BERT-Based & Development& 0.997& 0.970& 0.983& 0.980 &0.976& - \\
 &&   Validation & 0.989& 0.878& 0.893& 0.915& 0.876 &- \\
Sub-Section & Rule-Based & Development &- &0.970& 0.977 &0.989& 0.971 &-\\
 &&   Validation&  - &0.973& 0.946& 0.975& 0.958 &- \\
Full & BERT-Based & Development& 0.996& 0.991 &0.983 &0.980& 0.987& 0.990 \\
 &&   Validation  &0.990& 0.939& 0.898 &0.921& 0.911 &0.926 \\
    \bottomrule
  \end{tabular}
\end{table}

\end{document}